\documentclass[conference]{IEEEtran}
\IEEEoverridecommandlockouts

\def\BibTeX{{\rm B\kern-.05em{\sc i\kern-.025em b}\kern-.08em
    T\kern-.1667em\lower.7ex\hbox{E}\kern-.125emX}}

\title{WILD SAM: A Simulated-and-Real Data Augmentation for Autonomous Driving Perception under Challenging Weather

\thanks{© 2026 IEEE.  Personal use of this material is permitted.  Permission from IEEE must be obtained for all other uses, in any current or future media, including reprinting/republishing this material for advertising or promotional purposes, creating new collective works, for resale or redistribution to servers or lists, or reuse of any copyrighted component of this work in other works.}
}
\author{\IEEEauthorblockN{Hamed Khatounabadi}
\IEEEauthorblockA{\textit{Electrical and Computer Engineering} \\
\textit{Michigan State University}\\
East Lansing, USA \\
khatouna@msu.edu}
\and
\IEEEauthorblockN{Xiaohu Lu}
\IEEEauthorblockA{\textit{Electrical and Computer Engineering} \\
\textit{Michigan State University}\\
East Lansing, USA \\
luxiaohu@msu.edu}
\and
\IEEEauthorblockN{Hayder Radha}
\IEEEauthorblockA{\textit{Electrical and Computer Engineering} \\
\textit{Michigan State University}\\
East Lansing, USA \\
radha@msu.edu}
}

\usepackage{color, colortbl}
\usepackage{tasks}
\usepackage{multirow}
\usepackage{subfigure}
\usepackage{cite}
\usepackage{amsmath,amssymb,amsfonts}
\usepackage{algorithmic}
\usepackage{graphicx}
\usepackage{textcomp}
\usepackage{xcolor}
\usepackage[T1]{fontenc}
\usepackage{lmodern}
\usepackage{newtxtext,newtxmath}
\usepackage{hyperref}

\definecolor{greenmine}{rgb}{0.85, 0.85, 0.85}
\begin{document}
\maketitle
\begin{abstract}

The performance of state-of-the-art object detectors degrades significantly under adverse weather, causing a safety-critical domain shift problem for autonomous vehicles. Recent efforts address this problem by relying 
on synthetic data to train the object detectors, which limits their real-world applicability. Meanwhile, pseudo-labeling is widely used for cross-dataset domain adaptation problems.
However, these methods have not been exploited by weather-based domain adaptation approaches due to the noisy nature of such labels generated under harsh weather conditions. In this paper, we propose two new approaches to mitigate this weather-induced domain shift. First, we propose a Weather-Induced pseudo-Label Denoising (WILD) framework that filters noisy pseudo labels generated by real data captured under adverse weather conditions. Second, we develop a novel hybrid training methodology, WILD SAM, that exploits both pseudo-label denoising and simulation-based training solutions while using real-data from the target harsh-weather domain.
We validate both proposed approaches, WILD and WILD SAM, on the recently released Four Seasons dataset across rainy and snowy scenarios. 
Experiments show that the proposed frameworks improve Average Precision (AP) up to 13\% and significantly reduce the weather-induced performance gap relative to the baseline. The code is available at: \href{https://github.com/Kh-Hamed/WILD-SAM}{https://github.com/Kh-Hamed/WILD-SAM}

\end{abstract}


\section{Introduction}

It is well known that LiDAR-based 3D object detection \cite{lang2019pointpillars, yan2018second, deng2021voxel, shi2023pv} becomes challenging under adverse weather conditions such as rain, snow, and dense fog. These conditions degrade and attenuate LiDAR returns, causing detectors to generate unreliable predictions. Such inaccuracies raise serious safety risks, especially for vulnerable road users like pedestrians and cyclists.

In the area of 2D object detection, existing approaches typically rely on data augmentation or feature-level domain adaptation to address domain shifts. For instance, Robust Object Detection in Challenging Weather Conditions \cite{gupta2024robust} proposes synthetic image generation using Cyclic-GAN  \cite{zhu2017unpaired} and style-transfer\cite{gatys2015neural} to mimic harsh weather scenarios. MS-DAYOLO \cite{hnewa2023integrated} is another example that employs multi-scale, multi-level adversarial layers to learn domain-invariant features, while Domain Adaptive Faster R-CNN \cite{DA-Faster-RCNN-8578450} reduces domain shift at both global and instance levels using gradient-reversal layers. Although these approaches are effective in 2D, their direct applicability to 3D detection is limited because of the sparse nature of LiDAR point clouds.

\begin{figure}[t]
    \centering
    \includegraphics[width=1\linewidth]{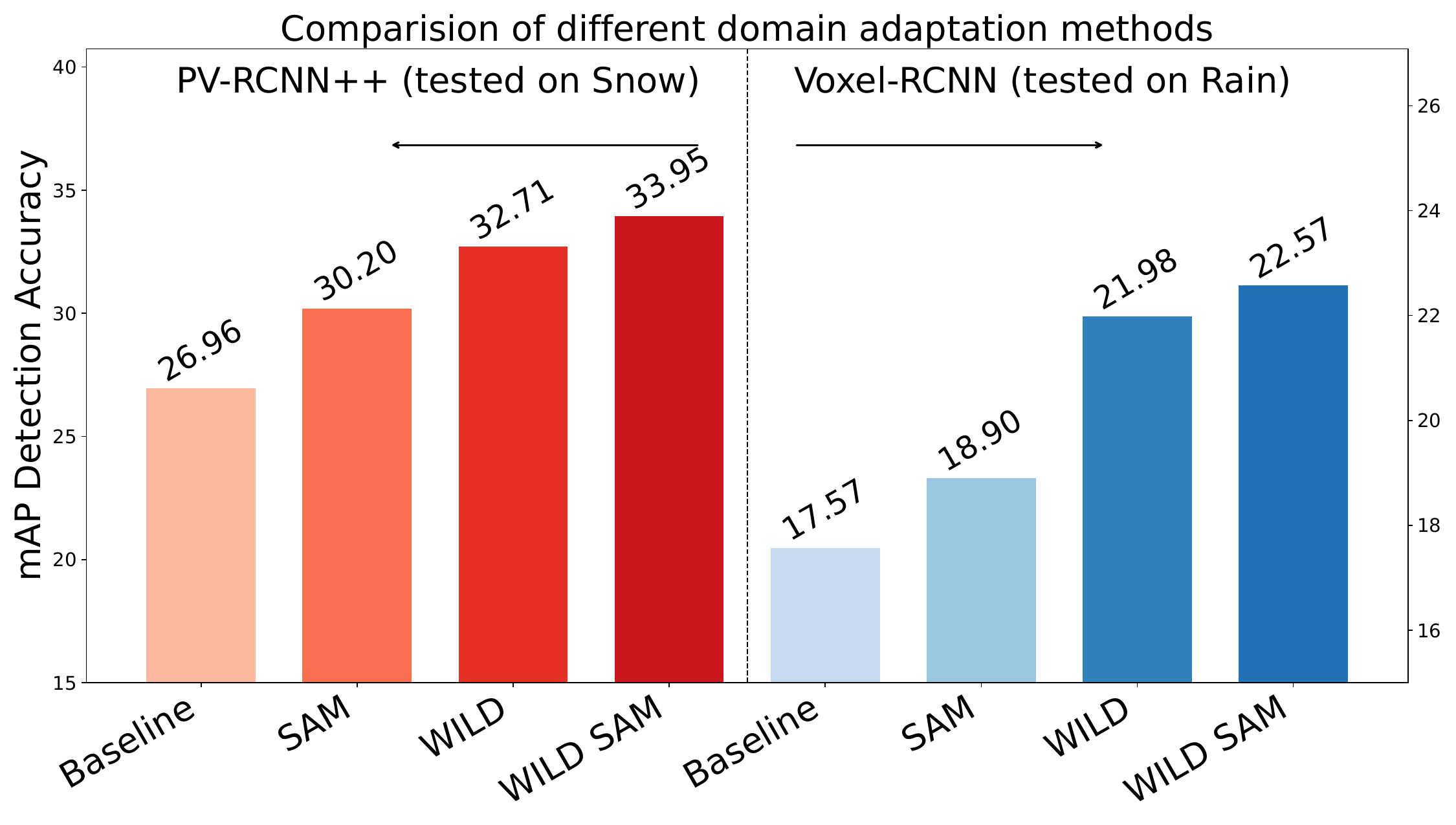}
    \caption{Performance of different variants of domain adaptation frameworks in comparison with baseline 3D object detector models when tested under both rainy and snowy conditions. The bar chart reports mean Average Precision (mAP \protect\footnotemark[1]) under four training scenarios: Baseline (trained only on summer data), Physics-based Simulated Augmentation and Modeling (SAM) frames, Weather-Induced Pseudo-Label Denoising (WILD) Augmented frames, and WILD SAM. While SAM frames yield some improvements, the proposed WILD SAM approaches outperform the simulation-based strategy consistently, and they increase mAP for both LiDAR 3D detectors (PV-RCNN++ at left, and Voxel-RCNN at right). Results are evaluated on the MSU-4S dataset \cite{kent2024msu}.}
    \label{fig:teaser}
\end{figure}
\footnotetext[1]{mAP is computed with IoU thresholds of 0.7 for Car, 0.5 for Pedestrian, and 0.25 for Bike.}

Under 3D detection, many recent studies focus on closing the domain gap across datasets. Approaches such as adaptation via proxy \cite{li2023adaptation}, DTS \cite{hu2023density}, SN \cite{wang2020train}, STD \cite{yang2021st3d}, and STD++ \cite{yang2022st3d++} primarily address annotation and scale mismatches between datasets. For example, STD~\cite{yang2021st3d} proposes Random Object Scaling (ROS) to address annotation–domain mismatch and introduces a Quality-aware Triplet Memory Bank (QTMB) to reduce pseudo-label noise during iterative refinement. DALI~\cite{DALI} proposes post-training size normalization (PTSN) to align average object sizes across domains and proposes Ray-Constrained Pseudo Point-Cloud Generation (RC-PPCG) to align the point with pseudo-labels such that they match the true geometric distribution of objects. While these methods effectively reduce gaps arising from annotation and scale differences, their applicability is limited when faced with weather-induced shifts that occur within a dataset.

Recent domain adaptation methods for object detection that target adverse-weather domain shift typically follow one of two strategies. The first strategy increases architectural complexity by incorporating additional sensor modalities and feature fusion (feature-level adaptation): examples include Robust-FusionNet \cite{zhang2022robust} and L4DR \cite{huang2025l4dr}, which fuse LiDAR with RGB or radar to address sparse or noisy LiDAR returns. While multi-modal fusion can improve robustness, it usually imposes architectural constraints (additional sensors, heavier backbones, calibration requirements) and increases test-time complexity. The second strategy uses physics-based simulation to augment training data (data-level adaptation), and they have traditionally focused on LiDAR-based systems, which is the focus of our work. As summarized in Table \ref{tab:sim_methods_weather}, representative works in this category include LiDAR Snowfall Simulation \cite{hahner2022lidar} and LISA \cite{kilic2025lidar}, which model weather effects by modulating return intensities and sampling snowflake/occlusion processes. Simulation approaches avoid changing the detector architecture but suffer from being trained on synthetic data (only), and hence, they do not fully capture the complexity of real target-domain weather. Consequently, these approaches fail to exploit available real target data, which can be readily used in unsupervised domain adaptation settings despite the lack of ground truth labels.

\begin{figure*}
    \footnotesize
    \centering
    \includegraphics[width=0.89\linewidth]{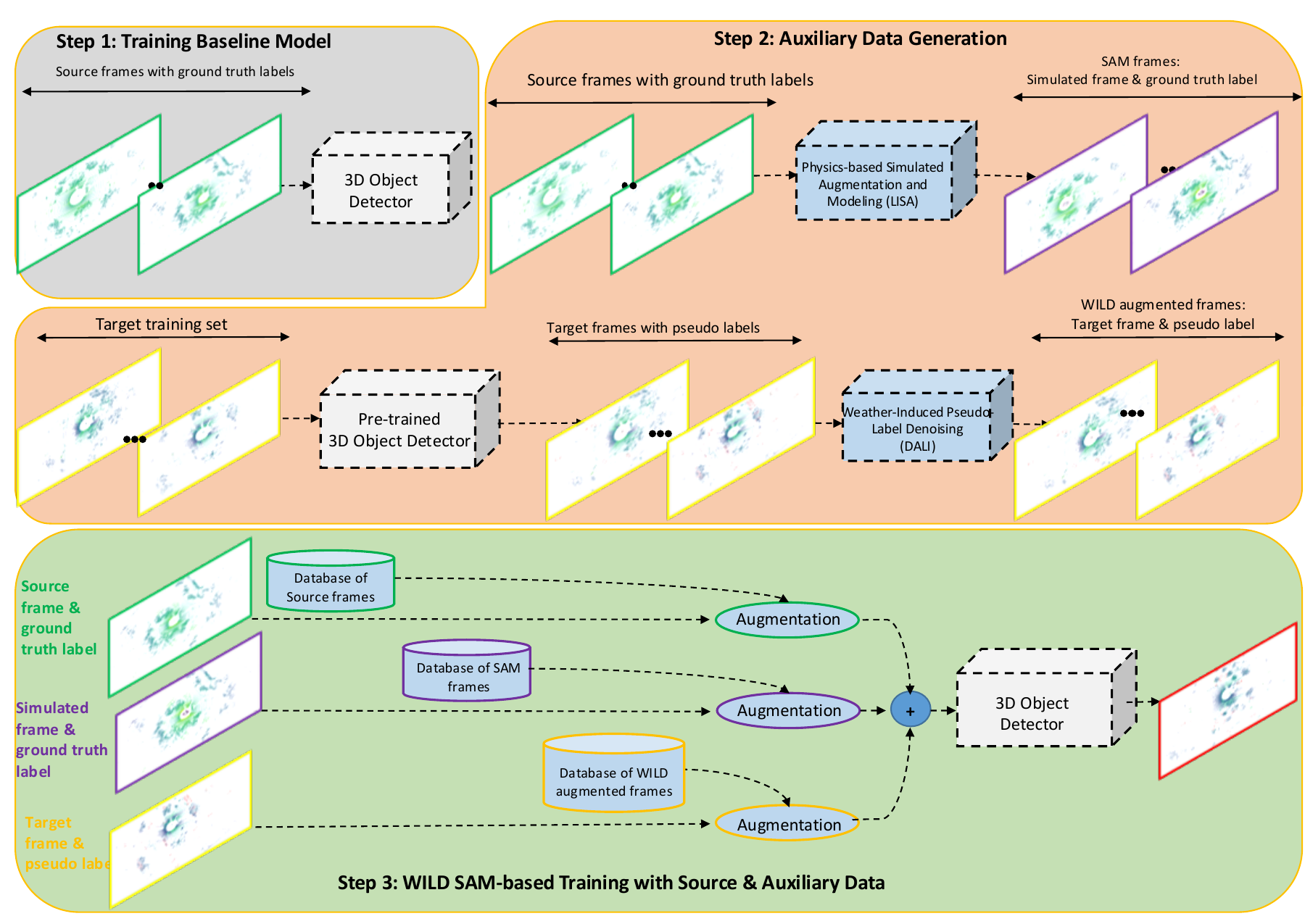}
    \caption{The proposed pipeline of WILD SAM-based training. \textbf{Step 1) Baseline training:} train an initial 3D object detector on labeled source-domain frames and their ground-truth annotations. \textbf{Step 2) Auxiliary data generation:} (top) apply a harsh-weather simulator to source frames to generate Physics-based Simulated Augmentation and Modeling (SAM) frames with the original ground-truth labels; (bottom) run the pre-trained 3D detector on unlabeled target-domain frames to produce pseudo labels and apply pseudo-label denoising to generate Weather-Induced Pseudo-Label Denoising (WILD) Augmented frames for training frames of the target domain. \textbf{Step 3) WILD SAM-based training with source and auxiliary data:} using three data banks (source frames with GT, simulated frames with GT, and target frames with denoised pseudo labels), we apply foreground object sampling augmentations, and jointly use these augmented sets to train the final 3D object detector for improved robustness under harsh weather.}
    \label{fig:arch_SAM}
\end{figure*}

To address these limitations, we propose a new hybrid strategy that exploits pseudo labeling denoising and simulated data augmentation. Despite their popularity for many unsupervised domain-adaptation problem areas, pseudo labeling-based techniques suffer from noisy labels, especially when generated by noisy data, such as ones captured under rainy or snowy conditions. Hence, our first objective is to develop a pseudo-label denoising approach that is tailored for harsh-weather data. Second, we aim at exploiting both simulated and real harsh-weather data to generate augmented frames for training object detectors. The effectiveness of this hybrid strategy is demonstrated under challenging weather conditions (see Fig.~\ref{fig:teaser}), with WILD SAM outperforming both pseudo-label denoising and simulation-based augmentation when applied alone. Therefore, we can summarize the key contributions of this paper as follows:
\begin{itemize}

\item We propose a Weather-Induced pseudo-Label Denoising (WILD) framework that filters noisy pseudo-labels generated by real data captured under adverse weather conditions. Hence, WILD exploits real data, and it enables the training of 3D object detectors with less noisy point cloud frames containing more reliable pseudo labels.
\item We develop a novel hybrid training methodology, WILD SAM, that exploits pseudo-label denoising and simulation-based training solutions while using real-data from the target harsh-weather domain. In particular, we employ WILD-based denoising in addition to a Simulated Augmentation and Modeling (SAM) of inclement weather to train an object detector. Hence, this overall WILD SAM augmentation strategy combines pseudo-label denoising with the utility of both real and simulated data to achieve a novel unsupervised domain adaptation framework.

\item We conducted multiple experiments with two popular 3D detectors, Voxel R-CNN and PV-RCNN++, on the MSU-4S dataset \cite{kent2024msu} to show the effectiveness of our method. Results show that both WILD and WILD SAM consistently improve robustness when tested on real data captured under adverse weather.
\end{itemize}

\begin{table}[!t]
  \caption{Simulation-based methods addressing weather-related domain shifts. Our method is a hybrid approach that employs target-domain real data and simulated data to adapt the detector.}
  \centering
  \resizebox{\columnwidth}{!}{
    \begin{tabular}{|c|c|c|}
      \hline
      \textbf{Method} & 
      \textbf{Target Weather} &
      \textbf{Modeling} \\
      \hline
      LiDAR Snowfall Simulation \cite{hahner2022lidar} 
        & Snow & Per-beam particle occlusion \\ \hline
      Fog Simulation for LiDAR \cite{hahner2021fog} 
        & Fog & Atmospheric scattering + attenuation  \\ \hline
      LISA \cite{kilic2025lidar} 
        & Rain, Snow, Fog & Hybrid Monte-Carlo and Mie scattering \\ \hline
      Cold-Weather 3D Detection \cite{piroli2022robust} 
        & Cold / Icing & Data augmentation + gas exhaust simulation \\ \hline
      \textbf{WILD SAM (Ours)} 
        & Rain, Snow & Simulation and real data\\ \hline
    \end{tabular}
  }
  \label{tab:sim_methods_weather}
\end{table}

\section{Methodology}

\begin{table*}[!t]
\caption{Improved detection accuracy of two popular detectors, PV-RCNN++ and Voxel-RCNN, under WILD SAM-based training. All experiments use the summer segment as the source domain. WILD SAM-snow denotes models trained with snow-based physics simulation and denoised pseudo-labels on the snowy target frames, while WILD SAM-rain denotes models trained with rainy counterparts. (a) shows results tested on snow, and (b) shows results tested on rain. Cross-weather evaluation (e.g., testing WILD SAM-snow on rain) highlights the generalization ability of WILD SAM.}
\label{tab:SAM-based-result-MSU-FS-rain-snow}
\centering
\subfigure[Tested on: Snow]{%
  \begin{minipage}[t]{0.496\textwidth}
    \centering
    \resizebox{\linewidth}{!}{%
      \begin{tabular}{|c|c|c|c|}
      \hline
      Detectors $\downarrow$ & Car & Pedestrian & Bike \\ 
              & AP\_R40@0.70 & AP\_R40@0.50 & AP\_R40@0.25 \\ 
      \hline
      Voxel-RCNN (baseline) \cite{deng2021voxel} 
          & 21.02 & 29.49 & 22.49 \\ 
      \rowcolor{greenmine}
      + WILD SAM-snow (\textit{$\Delta$}) 
          & +4.23 & +13.47 & +3.25 \\ 
      + WILD SAM-rain (\textit{$\Delta$}) 
          & +3.82 & +13.86 & \textbf{-3.52} \\ 
      \hline\hline
      PV-RCNN++ (baseline) \cite{shi2021pv} 
          & 23.46 & 42.68 & 14.73 \\ 
      \rowcolor{greenmine}
      + WILD SAM-snow  (\textit{$\Delta$})
          & +3.99 & +11.70 & +5.36 \\ 
      + WILD SAM-rain  (\textit{$\Delta$})
          & +3.14 & +10.94 & +0.74 \\ 
      \hline
      \end{tabular}
    }
    \label{tab:SAM MSU-FS snow}%
  \end{minipage}%
}%
\hfill
\subfigure[Tested on: Rain]{%
  \begin{minipage}[t]{0.496\textwidth}
    \centering
    \resizebox{\linewidth}{!}{%
      \begin{tabular}{|c|c|c|c|}
      \hline
      Detectors $\downarrow$ & Car & Pedestrian & Bike \\ 
              & AP\_R40@0.70 & AP\_R40@0.50 & AP\_R40@0.25 \\ 
      \hline
      Voxel-RCNN (baseline) \cite{deng2021voxel} 
          & 16.51 & 33.52 & 2.67 \\ 
      \rowcolor{greenmine}
      + WILD SAM-rain (\textit{$\Delta$})
          & +1.81 & +10.27 & +2.94 \\ 
      + WILD SAM-snow (\textit{$\Delta$})
          & +4.38 & +10.66 & +2.91 \\ 
      \hline\hline
      PV-RCNN++ (baseline) \cite{shi2021pv} 
          & 20.93 & 40.90 & 2.38 \\ 
      \rowcolor{greenmine}
      + WILD SAM-rain  (\textit{$\Delta$}) 
          & +1.07 & +7.70 & +3.69 \\ 
      + WILD SAM-snow  (\textit{$\Delta$}) 
          & +3.31 & +7.83 & +3.47 \\ 
      \hline
      \end{tabular}
    }
    \label{tab:SAM MSU-FS rain}%
  \end{minipage}%
}%
\end{table*}

\begin{table}[t]
    \caption{Detection accuracy of Voxel-RCNN and PV-RCNN++ evaluated on summer (source) and on snow/rain (target) domain. $\Delta$ rows show the domain shift (source $\rightarrow$ target) in Average Precision (AP).}
    \resizebox{\linewidth}{!}{%
        \begin{tabular}{|c|c|c|c|c|}
        \hline
        Detector & Test set & Car @0.70 & Pedestrian @0.50 & Bike @0.25 \\
        \hline
        Voxel-RCNN & Summer & 68.95 & 58.87 & 43.71 \\
        Voxel-RCNN & Snow   & 21.02 & 29.49 & 22.49  \\
        \rowcolor{greenmine}
        $\Delta$ (Summer - Snow)& - & \textbf{47.93} & \textbf{29.38} & \textbf{21.22} \\
        Voxel-RCNN & Rain   & 16.51 & 33.52 & 2.67  \\
        \rowcolor{greenmine}
        $\Delta$ (Summer - Rain)& - & \textbf{52.44} & \textbf{25.35} & \textbf{41.04} \\
        \hline
        PV-RCNN++  & Summer & 65.29 & 61.09 & 41.35 \\
        PV-RCNN++  & Snow   & 23.46 & 42.68 & 14.73  \\
        \rowcolor{greenmine}
        $\Delta$ (Summer - Snow)& - & \textbf{41.83} & \textbf{18.41} & \textbf{26.62} \\
        PV-RCNN++  & Rain   & 20.93 & 40.90 & 2.38  \\
        \rowcolor{greenmine}
        $\Delta$ (Summer - Rain)& - & \textbf{44.36} & \textbf{20.19} & \textbf{38.97} \\
        \hline
        \end{tabular}
    }
    \label{tab:summer-rain-snow-domain-shifts}
\end{table}


Under WILD SAM-based training, we augment the original source-domain dataset with: (a) weather-induced labeled data and corresponding frames, and (b) simulated frames that represent the target-domain weather to reduce the domain shift caused by harsh weather. Figure \ref{fig:arch_SAM} illustrates our proposed pipeline. Our framework follows an unsupervised domain adaptation (UDA) strategy where we have access to labeled source-domain frames and unlabeled target-domain frames. Thus, we train an initial 3D LiDAR-based detector on the source data (see step 1 in Figure \ref{fig:arch_SAM}) and use it in the next step to generate pseudo-labels for the target frames. In step 2, we build two additional, independent datasets (together with the original source set) and use all three sets jointly to retrain the WILD SAM-based detector. The procedures for generating these auxiliary datasets and the applied augmentations are described in detail in the following subsections.

\subsection{Auxiliary Data Generation}
\subsubsection{Weather-Induced Pseudo-Label Denoising (WILD) Augmented frames}
 
The main challenge in UDA is the absence of ground-truth labels in the target domain. To provide supervision for the target domain frames, we employ the model trained on the source domain to generate pseudo-labels for the target domain training set. However, these pseudo-labels are noisy and contain a significant number of false positives. 

To address the issue of pseudo-label noise, we adapt the ray-constrained pseudo point-cloud generation (RC-PPCG) method from DALI \cite{DALI} to handle harsh weather target domain data. The RC-PPCG framework denoises pseudo-labels by projecting points within target-domain bounding boxes onto a dense reference model (e.g., a CAD model or high-density source point cloud) via ray tracing.
Following the point-based strategy described in \cite{DALI}, we construct our reference library using dense lidar samples extracted from source-domain (summer segment) ground truth boxes for each class. In our implementation, we replace the target points' $(x, y, z)$ coordinates with the projected surface locations while preserving their original lidar intensities. We modify points in pseudo-labels with fewer than N points, which are likely noisy and unreliable. This ensures that, even if a false positive pseudo-label mistakenly corresponds to a non-object region (e.g., a bush), the associated points are rearranged to follow the geometry distribution of the intended target object class (e.g., a car).

It is worth noting that ground-truth database sampling is a standard augmentation strategy for 3D object detectors. In this process, a database is constructed from ground-truth bounding boxes and their associated points, which are then sampled to augment the source frames. Following a similar strategy, we construct an independent database from the denoised pseudo-labels of the target frames by rearranging the points within each pseudo-label. This target-domain database is then used in WILD frames training separately from the ground-truth database of the source domain.

\subsubsection{Physics-based Simulated Augmentation and Modeling (SAM) frames}

We use physics-based simulation to further diversify the training data. In particular, we employ LISA \cite{kilic2025lidar} to synthesize harsh-weather conditions (e.g., snow and rain). For each labeled source frame, a corresponding simulated version is generated using LISA, where the weather intensity is controlled by the precipitation-rate parameter $\tau$ (mm/hr). LISA randomly places particles throughout the scene. For each particle, it checks whether the laser pulse backscattered by the particle is stronger than the pulse reflected by the original point (e.g., from a car). If the particle’s backscatter dominates, the original return is occluded and the corresponding LiDAR point is lost; if the actual point's reflection is stronger, the point remains, but its intensity is attenuated. The point remains unchanged if a particle does not lie on the laser ray that hits it. So, original ground-truth annotations remain valid and are kept aligned with the simulated frames. As described in the previous subsection, we construct a separate synthetic ground-truth database from these simulated frames, maintained independently of the original source ground-truth database and Weather-Induced Pseudo-Label Denoising (WILD) database (see the independent databases illustrated in Step 3 of Figure \ref{fig:arch_SAM}).

\subsection{WILD SAM Training}
Under WILD SAM, we retrain the 3D object detector using three independent sets of lidar point-cloud frames: (a) source-domain frames with their ground-truth labels, (b) augmented target domain frames with denoised pseudo labels using Weather-Induced pseudo-Label Denoising (WILD) , and (c) Physics-based Simulated Augmentation and Modeling (SAM) frames with source-domain ground-truth labels. To maximize input diversity, each dataset using its own ground-truth / pseudo-label database is further augmented independently. After adding additional foreground objects, we apply augmentations such as random flips (along the x and/or y axis) and random rotations to improve the model’s generalization one step further. As illustrated in Figure \ref{fig:arch_SAM}, this setup triples the training batch size under the full WILD SAM framework, while the model architecture and test-time inference cost remain unchanged.

\section{Experiments}

\begin{figure*}[t]
    \footnotesize
    \centering
    \includegraphics[width=0.89\linewidth]{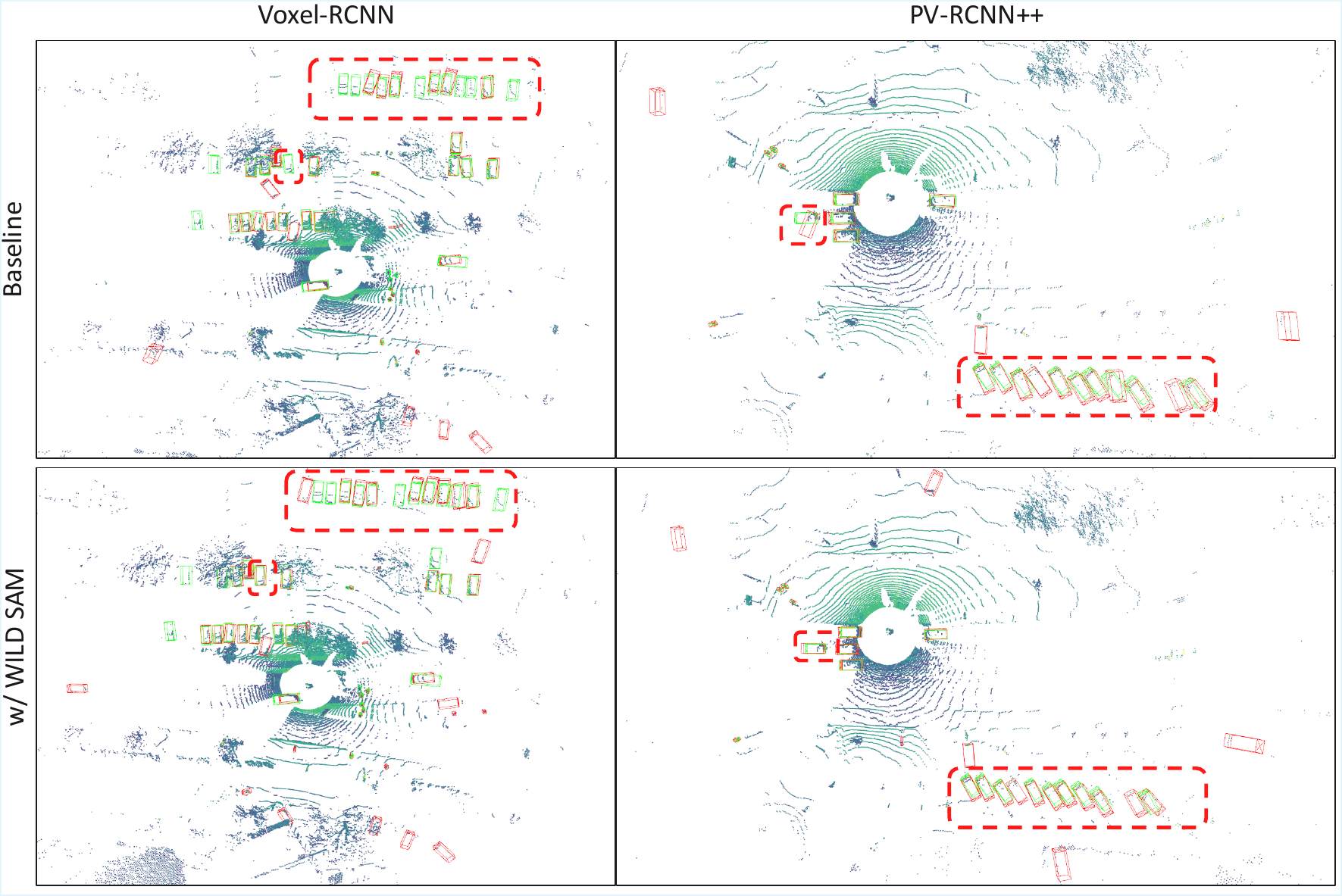}
    \caption{Visual examples of the proposed WILD SAM method in improving the robustness of 3D object detectors under harsh weather conditions. In each column, the top row shows baseline predictions, while the second row shows predictions under WILD SAM-based training. The left column presents Voxel-RCNN predictions under snow, and the right column shows PV-RCNN++ predictions under rain. WILD SAM effectively reduces the false negative rate.}
    \label{fig: SAM visual examples}
\end{figure*}
\begin{table*}[!t]
    \caption{Ablation study of WILD SAM components on PV-RCNN++ and Voxel-RCNN using the MSU-4S dataset. Results show that physics-based Simulated Augmentation and Modeling (SAM) alone may degrade detection accuracy under rain while remaining effective in snow. In contrast, Weather-Induced Pseudo-Label Denoising (WILD) augmented frames improve detection accuracy under both rain and snow; combining both components can achieve the highest overall mAP gain.}
    \centering
    \subfigure[Tested on: Rain]{%
      \begin{minipage}[t]{0.496\textwidth}
        \centering
        \resizebox{\linewidth}{!}{
        \begin{tabular}{|c|c|c|c|c|c|c|}
            \hline
            \multirow{2}{*}{\textbf{PV-RCNN++ \cite{shi2023pv}}} & 
            \multirow{2}{*}{\textbf{SAM}} & 
            \multirow{2}{*}{\textbf{WILD}} & 
            \textbf{Car} & 
            \textbf{Pedestrian} & 
            \textbf{Bike} &
            \textbf{mAP} \\
            \cline{4-7}
             & & & 
             AP\_R40 @0.7 & AP\_R40 @0.5 & AP\_R40 @0.25 &  \\
            \hline
            Baseline & – & – & 20.93 & 40.90 & 2.38 & 21.40 \\ 
            \hline
            & \checkmark & – & 18.53 & 39.20 & 2.85 & 20.19 \\
            \rowcolor{greenmine}
            $\Delta$ & – & – & -2.40 & -1.70 & +0.47 & -1.21 \\
            \hline
            & – & \checkmark & 22.37 & 48.79 & 5.11 & 25.42 \\
            \rowcolor{greenmine}
            $\Delta$ & – & – & \textbf{+1.44} & \textbf{+7.89} & +2.73 & +4.02 \\
            \hline
            & \checkmark & \checkmark & 21.99 & 48.60 & 6.07 & 25.55 \\
            \rowcolor{greenmine}
            $\Delta$ & – & – & +1.06 & +7.70 & \textbf{+3.68} & \textbf{+4.15} \\
            \hline
        \end{tabular}
        }
        \label{tab: ablation SAM pv-rcnn rain}%
      \end{minipage}%
    }%
    \hfill
    \subfigure[Tested on: Snow]{%
      \begin{minipage}[t]{0.496\textwidth}
        \centering
        \resizebox{\linewidth}{!}{
        \begin{tabular}{|c|c|c|c|c|c|c|}
            \hline
            \multirow{2}{*}{\textbf{Voxel-RCNN \cite{deng2021voxel}}} & 
            \multirow{2}{*}{\textbf{SAM}} & 
            \multirow{2}{*}{\textbf{WILD}} & 
            \textbf{Car} & 
            \textbf{Pedestrian} & 
            \textbf{Bike} &
            \textbf{mAP} \\
            \cline{4-7}
             & & & 
             AP\_R40 @0.7 & AP\_R40 @0.5 & AP\_R40 @0.25 &  \\
            \hline
            Baseline & – & – & 21.02 & 29.49 & 22.49 & 24.33 \\ 
            \hline
            & \checkmark & – & 23.90 & 43.95 & 28.05 & 31.97 \\
            \rowcolor{greenmine}
            $\Delta$ & – & – & +2.88 & \textbf{+14.46} & \textbf{+5.56} & \textbf{+7.64} \\
            \hline
            & – & \checkmark & 24.98 & 42.69 & 23.33 & 30.33 \\
            \rowcolor{greenmine}
            $\Delta$ & – & – & +3.96 & +13.20 & +0.84 & +6.00 \\
            \hline
            & \checkmark & \checkmark & 25.25 & 42.96 & 25.74 & 31.32 \\
            \rowcolor{greenmine}
            $\Delta$ & – & – & +\textbf{4.23} & +13.47 & +3.25 & +6.99 \\
            \hline
        \end{tabular}
        }
        \label{tab: ablation SAM voxel-rcnn snow}%
      \end{minipage}%
    }%

    \label{tab: ablation SAM both}
\end{table*}


\begin{table}[!t]
    \caption{Effect of varying precipitation rate on Voxel-RCNN performance when adapting from Summer to Snow using only Physics-based Simulated Augmentation and Modeling (SAM) frames.}
    \centering
    \resizebox{0.95\linewidth}{!}{
    \begin{tabular}{|c|c|c|c|c|c|}
        \hline
        \multirow{2}{*}{\textbf{Voxel-RCNN \cite{deng2021voxel}}} & 
        \multirow{2}{*}{\textbf{Rate (mm/hr)}} & 
        \textbf{Car} & 
        \textbf{Pedestrian} & 
        \textbf{Bike} &
        \textbf{mAP} \\
        \cline{3-6}
         & & 
         AP\_R40 @0.7 & AP\_R40 @0.5 & AP\_R40 @0.25 &  \\
        \hline
        Baseline (Summer) & – & 21.02 & 29.49 & 22.49 & 24.33 \\ 
        \hline
        +SAM frames  & 5 mm/hr & 23.90 & 43.95 & 28.05 & 31.97 \\
        \rowcolor{greenmine}
        $\Delta$ & – & +2.88 & +14.46 & +5.56 & +7.64 \\
        \hline
        +SAM frames  & 10 mm/hr & 23.86 & 44.35 & 29.39 & 32.53 \\
        \rowcolor{greenmine}
        $\Delta$ & – & +2.84 & +14.86 & +6.90 & +8.20 \\
        \hline
        +SAM frames & 20 mm/hr & 24.18 & 42.33 & 25.25 & 30.59 \\
        \rowcolor{greenmine}
        $\Delta$ & – & +3.16 & +12.84 & +2.76 & +6.26 \\
        \hline
    \end{tabular}
    }

    \label{tab: voxel-rcnn-snow-rate}
\end{table}


\subsection{Dataset}
We conducted all experiments on the MSU-4S dataset, which contains data collected under different weather conditions, including sunny, rainy, and snowy. Each annotated weather segment contains approximately 3k frames for training and 2k frames for testing. For all experiments, we used four input features (x, y, z, i), where "x", "y", "z" are point coordinates, and "i" is the lidar intensity. The summer (sunny) segment was designated as the source domain, whereas the snowy and rainy segments were considered target domains.

\subsection{Hyperparameters}
For all experiments on the MSU-4S dataset, we trained each model for 80 epochs using the Adam optimizer with a one-cycle scheduler and a learning rate of 0.01. All models were implemented based on the OpenPCDet \cite{openpcdet2020}. For the LISA rain and snow simulation, precipitation was synthesized with $\tau=5 mm/hr$. We set N to identify noisy pseudo-labels for WILD frames as 50, 20, and 20 for cars, pedestrians, and bikes, respectively.

\subsection{Results and Discussion}
We conducted multiple experiments to demonstrate the presence of a clear domain shift under harsh weather conditions. As shown in Table~\ref{tab:summer-rain-snow-domain-shifts}, adverse weather significantly impacts detection performance. As shown, models trained on clear-weather (summer) frames exhibit a substantial drop in accuracy when evaluated on rainy or snowy conditions. This highlights a clear geometric and intensity domain gap, which we aim to address using WILD SAM-based training.

Table~\ref{tab:SAM-based-result-MSU-FS-rain-snow} shows the effectiveness of WILD SAM-based training to close this domain gap for Voxel-RCNN and PV-RCNN++. Our proposed WILD SAM framework consistently improves detection accuracy under harsh weather conditions by more than 13\%. All models were trained on the summer split (source domain) and evaluated on snow/rain scenarios (target domain). Furthermore, Table~\ref{tab:SAM-based-result-MSU-FS-rain-snow} demonstrates that WILD SAM provides significant gains for vulnerable road users such as pedestrians and cyclists, thereby addressing critical safety concerns in adverse weather conditions. Moreover, the WILD SAM-snow model, trained with Snow Physics-based Harsh Weather simulated frames and snow Denoised pseudo Label frames, demonstrates robustness and has even been tested under rainy conditions, as well as the WILD SAM-rain models. This suggests that snowy and rainy conditions share similar characteristics, enabling WILD SAM-trained models to generalize effectively across different types of harsh weather, such as rain and snow.

\subsection{Ablation Studies}
We evaluated the contributions of SAM and WILD augmented frames under WILD SAM-based training. Table \ref{tab: ablation SAM both} reports results for both PV-RCNN++ and Voxel-RCNN. The results show that most of the detection gain comes from the WILD augmented frames; synthetic SAM data provides additional improvements, but cannot fully model the true target-domain distribution.

Table \ref{tab: ablation SAM pv-rcnn rain} shows that using synthetic data alone can sometimes slightly hurt performance, whereas combining synthetic samples with WILD augmented frames  can lead to the highest overall increase in mAP. Conversely, Table \ref{tab: ablation SAM voxel-rcnn snow} demonstrates that for certain weather conditions (e.g., snow here), synthetic data can be as beneficial as WILD frames. 

Overall, Table \ref{tab: ablation SAM both} highlights that precise point–box matching plays a critical role in achieving robust detection accuracy improvement for different weather conditions. While synthetic data enhances diversity and provides more target-like samples, it may still fall short of accurately representing the true target distribution under specific weather conditions.

We further analyzed the effect of varying the precipitation rate parameter $\tau$, which controls snow intensity in the LISA simulation. Table~\ref{tab: voxel-rcnn-snow-rate} summarizes results when adapting from summer to snowy conditions using only SAM frames. As observed, increasing $\tau$ generally incrementally improves mAP up to a certain level, after which detection accuracy gains begin to diminish. To maintain consistency across experiments and avoid incremental fine-tuning for each weather condition and detector, we set $\tau = 5$~mm/hr for all WILD SAM experiments.

\subsection{Visual Examples}
Figure \ref{fig: SAM visual examples} presents visual examples demonstrating the behavior of WILD SAM-based models under challenging weather conditions. Compared to the baseline, WILD SAM-based training increases recall by detecting more true objects (green boxes), though at the cost of introducing additional false positives (red boxes). In other words, the positive impact of reducing false negatives dominates, which is especially important for safety-critical applications such as autonomous driving. Although this leads to a slightly higher number of predictions, it enables better detection of previously missed objects without significantly compromising precision.

\section{Conclusion}
In this study, we demonstrated the domain shift caused by adverse weather conditions for two widely used LiDAR-based 3D detectors, PV-RCNN++ and Voxel-RCNN. To address this domain gap, we propose WILD SAM, a hybrid approach that integrates physics-based simulation with pseudo-label denoising within an Unsupervised Domain Adaptation (UDA) framework to enhance the robustness of LiDAR-based detectors. Our experiments on the MSU-4S dataset validate that WILD SAM-based training effectively mitigates domain shift while maintaining the same computational complexity during inference.

\noindent
\textbf{Acknowledgment} This work was supported by a Department of Transportation/Federal Railroad Administration grant.

\bibliographystyle{IEEEtran}
\bibliography{main}

\end{document}